\documentclass[runningheads]{llncs}

\usepackage{amssymb,amsmath,mathtools}
\usepackage{nicefrac}
\usepackage{mathrsfs}
\usepackage{xspace}
\usepackage[inline]{enumitem}
\usepackage{comment}
\usepackage{thm-restate}
\usepackage{graphicx}
\usepackage[colorlinks=true,linkcolor=violet,citecolor=blue]{hyperref}
\usepackage[capitalize,nameinlink]{cleveref}
\usepackage{adjustbox}
\usepackage{bm}
\usepackage{tikz}
\usepackage{ifthen}
\usetikzlibrary {arrows.meta}
\usetikzlibrary{backgrounds}
\usetikzlibrary{arrows,shapes}
\usetikzlibrary{tikzmark}
\usetikzlibrary{calc}
\usepackage{utfsym}
\usepackage[numbers]{natbib}
\usepackage{bussproofs}
\usepackage{latexsym} 
\usepackage{tcolorbox}
\usepackage{algorithm}
\usepackage{algpseudocode}
\usepackage{booktabs}
\usepackage{caption}
\usepackage{wrapfig}
\usepackage{makecell}
\usepackage{multirow}
\usepackage{bbding}
\spnewtheorem{subproblem}{Problem~1.\!\!}{\itshape}{}

\Crefname{problem}{Problem}{Problem}
\Crefname{subproblem}{Problem~1.\!\!}{Problem~1.\!\!}
\crefname{lemma}{Lem.}{Lem.}
\crefname{example}{Exmp.}{Exmp.}
\crefname{section}{Sect.}{Sect.}
\Crefname{appendix}{Appx.}{Appx.}
\crefname{definition}{Def.}{Def.}
\crefname{theorem}{Thm.}{Thm.}
\crefname{corollary}{Cor.}{Cor.}
\crefname{algorithm}{Alg.}{Alg.}

\usepackage{graphicx}
\usepackage{subcaption}

\usepackage{svg}
\usepackage{amsmath}
\usepackage[normalem]{ulem}
\usepackage{hhline}

\newcommand{\norm}[1]{\left\lVert#1\right\rVert}
\newcommand{\sfcommentinline}[1]{}

\newcommand{\oomit}[1]{}

\newcommand{\reviewercomment}[1]{}

\newcommand{\hscomment}[1]{}
\newcommand{\hscommentinline}[1]{}

\newcommand{\nzcomment}[1]{}
\newcommand{\nzcommentinline}[1]{}

\definecolor{myred}{HTML}{e41a1c}
\definecolor{myblue}{HTML}{377eb8}
\definecolor{mygreen}{HTML}{4daf4a}
\definecolor{mygray}{gray}{0.6}
\definecolor{myorange}{HTML}{ffa510}

\newcommand{\xx}{\bm{x}\xspace}

\makeatletter
\newsavebox{\@brx}
\newcommand{\llangle}[1][]{\savebox{\@brx}{\(\m@th{#1[}\)}%
  \mathopen{\copy\@brx\kern-0.6\wd\@brx\usebox{\@brx}}}
\newcommand{\rrangle}[1][]{\savebox{\@brx}{\(\m@th{#1]}\)}%
  \mathclose{\copy\@brx\kern-0.6\wd\@brx\usebox{\@brx}}}
\makeatother

\newcommand{\Init}[1][]{\ensuremath{\mathtt{Init}\ifthenelse{\equal{#1}{}}{}{(#1)}}\xspace}
\newcommand{\Inv}[1][]{\ensuremath{\mathtt{Inv}\ifthenelse{\equal{#1}{}}{}{(#1)}}\xspace}
\newcommand{\Flow}[1][]{\ensuremath{\mathtt{Flow}\ifthenelse{\equal{#1}{}}{}{(#1)}}\xspace}
\newcommand{\Jump}[1][]{\ensuremath{\mathtt{Jump}\ifthenelse{\equal{#1}{}}{}{(#1)}}\xspace}
\newcommand{\tJump}[1][]{\ensuremath{\mathtt{Jump}_t\ifthenelse{\equal{#1}{}}{}{(#1)}}\xspace}
\newcommand{\Dom}[1][]{\ensuremath{\mathtt{Dom}\ifthenelse{\equal{#1}{}}{}{(#1)}}\xspace}

\newcommand{\Traj}[1][]{\ensuremath{\xx\ifthenelse{\equal{#1}{}}{}{(#1)}}\xspace}
\newcommand{\dTraj}[1][]{\ensuremath{\dot{\xx}\ifthenelse{\equal{#1}{}}{}{(#1)}}\xspace}
\newcommand{\pTraj}[1][]{\ensuremath{\xx'\ifthenelse{\equal{#1}{}}{}{(#1)}}\xspace}

\makeatletter
\newcommand{\Pminus}{\mathbin{\mathpalette\prc@inner\relax}} 
\newcommand{\prc@inner}[2]{%
  \vbox{\offinterlineskip\m@th
    \ialign{%
      ##\cr
      \hidewidth\raisebox{-1.5\height}[0pt][0pt]{$#1.$}\hidewidth\cr
      $#1-$\cr
    }%
  }%
}
\makeatother

\newcommand{\until}[1][]{\ensuremath{\,\mathcal{U}\ifthenelse{\equal{#1}{}}{}{\ifthenelse{\pdfmatch{,}{#1}=1}{_{[#1]}}{_{#1}}}}\xspace}
\newcommand{\always}[1][]{\ensuremath{\Box\ifthenelse{\equal{#1}{}}{}{\ifthenelse{\pdfmatch{,}{#1}=1}{_{[#1]}}{_{#1}}}}\xspace}
\newcommand{\eventually}[1][]{\ensuremath{\Diamond\ifthenelse{\equal{#1}{}}{}{\ifthenelse{\pdfmatch{,}{#1}=1}{_{[#1]}}{_{#1}}}}\xspace}

\def\triangleforqed{\hbox{$\lhd$}}
\makeatletter
\DeclareRobustCommand{\qedT}{%
	\ifmmode
	\eqno \def\@badmath{$$}
	\let\eqno\relax \let\leqno\relax \let\veqno\relax
	\hbox{\triangleforqed}%
	\else
	\leavevmode\unskip\penalty9999 \hbox{}\nobreak\hfill
	\quad\hbox{\triangleforqed}%
	\fi
}
\makeatother

\titlerunning{Runtime Verification of POLAR-EXPRESS}

\begin{document}
\title{Case Study: Runtime Safety Verification of Neural Network Controlled System
    \thanks{Frank Yang, Simon Sinong Zhan, Yixuan Wang, and Qi Zhu’s work is partially supported by US National Science Foundation
grants 2324936 and 2328973. Chao Huang's work is supported by the grant EP/Y002644/1 under the EPSRC ECR International Collaboration Grants program, funded by the International Science Partnerships Fund (ISPF) and the UK Research and Innovation. }}

\author{Frank Yang\inst{1}
    \and
    Sinong Simon Zhan\inst{1} 
    \and
    Yixuan Wang\inst{1}
    \and
    Chao Huang\inst{2}
    \and 
    Qi Zhu\inst{1}
    }
    \authorrunning{F.~Yang et al.}

    \institute{Electrical and Computer Engineering, Northwestern University, Evanston, USA 
    \email{\{frankyang2024,SinongZhan2028,yixuanwang2024,qzhu\}@u.northwestern.edu}
    \and 
    School of Electronics and Computer Science, University of Southampton, Southampton, UK
    \email{Chao.Huang@soton.ac.uk}\\
    }

\maketitle              
\begin{abstract}
Neural networks are increasingly used in safety-critical applications such as robotics and autonomous vehicles. However, the deployment of neural-network-controlled systems (NNCSs) raises significant safety concerns. Many recent advances overlook critical aspects of verifying control and ensuring safety in real-time scenarios. This paper presents a case study on using POLAR-Express, a state-of-the-art NNCS reachability analysis tool, for runtime safety verification in a Turtlebot navigation system using LiDAR. The Turtlebot, equipped with a neural network controller for steering, operates in a complex environment with obstacles. We developed a safe online controller switching strategy that switches between the original NNCS controller and an obstacle avoidance controller based on the verification results. Our experiments, conducted in a ROS2 Flatland simulation environment, explore the capabilities and limitations of using POLAR-Express for runtime verification and demonstrate the effectiveness of our switching strategy. 
\end{abstract}

\section{Introduction}

The increasing complexity of control strategies used in cyber-physical systems (CPSs)~\cite{lee2016introduction}, specifically those based on neural networks, has revolutionized decision-making and control in several critical domains, including healthcare~\cite{xue2024prescribing,xue2023assisting}, robotics~\cite{wang2023joint,wang2023enforcing,zhan2023state}, transportation~\cite{deka2018transportation,liu2022physics,xiong2015cyber}, building control~\cite{wei2017deep,xu2022accelerate,xu2020one}, and industrial automation~\cite{breivold2015internet,wollschlaeger2017future}. These advanced control approaches excel at handling complex and dynamic environments due to their ability to learn and adapt from data. However, assuring the safety and stability of these systems for the nonlinearity of control systems and their closed-loop formation with dynamic systems remains a significant challenge ~\cite{alur1995algorithmic, sastry2013nonlinear,zhu2021safety,zhu2020know}.

Literature in this domain primarily focuses on developing methodologies to assess and guarantee the reliability and robustness of neural network decisions. Early approaches often relied on static analysis techniques that scrutinized network structures and weights to predict behavior under various inputs~\cite{albarghouthi2021introduction,huang2017safety,wang2021beta}. Recent advancements have introduced more dynamic methods, such as formal verification and reachability analysis~\cite{althoff2015introduction,dutta2019reachability,lopez2023nnv,wang2023polar}, which offer more nuanced insights into network behaviors across potential operational scenarios. The Simplex-based Reachability Analysis \cite{bak2016rtss, Desai2019SOTER} guarantees system overall safety by integrating a verified safety controller and decision logic that switches between complex and safety controllers. While making significant contributions on real-time reachability, the use of these verification tools in realistic environments with machine-learning components remains largely unexplored. Our work uses POLAR-Express \cite{wang2023polar}, a state-of-the-art verification tool to perform online reachability analysis for realistic robotic systems with neural network control and LiDAR sensing. We demonstrate the effectiveness of POLAR-Express for online reachability analysis by safely navigating robots in complex environments with obstacle constraints. The main contributions of this paper are as follows:
\begin{itemize}
    \item We present a comprehensive study demonstrating the feasibility of performing runtime verification by POLAR-Express on Turtlebot for safe navigation.
    \item We provide a safe online controller switching strategy to avoid unknown obstacles based on the runtime verification result.
\end{itemize}



\vspace{-10pt}
\section{Related Work}

\textbf{Runtime Verification for Control.}
Runtime Verification (RV) plays a crucial role in the real-time operation of autonomous systems, such as autonomous vehicles~\cite{huang2014rosrv}, transportation networks~\cite{qian2022formal} and medical devices~\cite{leucker2016runtime}. In control theory, techniques such as adaptive control and robust control are employed to manage uncertainties and ensure stability in real-time scenarios~\cite{astrom2010feedback,zhou1996robust}. From the formal methods perspective, model checking, which utilizes temporal logic specifications like linear temporal logic (LTL) and signal temporal logic (STL), forms the backbone of verification processes on the system's trajectories~\cite{bauer2011runtime,desai2017combining,havelund2019extension,jakvsic2018algebraic,su2024switching,zapridou2020runtime,zhang2023online}. Moreover, the integration of stochastic quantification tools with temporal logic through conformal prediction frameworks offers a formal statistical guarantee of system reliability under dynamic conditions~\cite{cairoli2023learning,lindemann2023conformal}. These developments have fostered innovative hybrid approaches that combine the strengths of control theory and formal methods to tackle complex verification challenges in real-time systems~\cite{alur2015principles,tabuada2009verification}.

\smallskip
\noindent
\textbf{NNCS Verification.}
The verification of neural networks has emerged as a critical research area~\cite{zhu2023verification}. Tools like Reluplex~\cite{katz2017reluplex}, Marabou~\cite{katz2019marabou}, and Sherlock~\cite{dutta2019sherlock} employ techniques derived from formal methods to ensure that neural networks adhere to specified safety and performance criteria. Others methods includes optimization-based over-approximation~\cite{dutta2019sherlock,wang2021beta}, and hybrid system approximation~\cite{ivanov2019verisig}. Alongside these, with the existing techniques for verifying dynamic systems~\cite{chen2013flow,chutinan2003computational}, a series of works have attained considerable maturity, providing formal analysis for neural network controlled systems (NNCSs)~\cite{althoff2015introduction,  dutta2019reachability,fan2020reachnn,fan2019towards,huang2019reachnn,kochdumper2023constrained,liu2021CORA,wang2022design,wang2023polar,wang2021bounding,wang2022efficient}. However, most of these approaches have not demonstrated the capability to verify NNCSs in a runtime environment. \cite{ivanov2020case} presented the feasibility of Verisig~\cite{ivanov2021verisig,ivanov2019verisig} using high-dimensional LiDAR measurements as the NNCS input, albeit in a simplistic setting for runtime requirements.

\vspace{-10pt}
\section{POLAR-Express Case Study}

\subsection{Preliminary}

\noindent
\textbf{NNCS.} We consider the explicit dynamics of an NNCS as $\dot{s} = f(s, a)$ where the state variable is $s \in \mathcal{S} \subseteq \mathbb{R}^n$, control input is $a \in \mathcal{A} \subseteq \mathbb{R}^m$, and the dynamic $f: \mathbb{R}^n \times \mathbb{R}^m \rightarrow \mathbb{R}^n$ is a Lipschitz continuous function, ensuring a unique solution of the ODE. Such a system can be controlled by a feedback NN controller $\kappa_{nn}$, at $i$-th ($i = 0, 1, \cdots$) sampling period $i\delta$, $\kappa_{nn}$ reads the system state $s_{i\delta}$, generates a control input $a = \kappa_{nn}(s_{i\delta})$, and the system evolves according to $\dot{s} = f(s, a)$ within the period of $[i\delta, (i+1)\delta]$. The \textit{flowmap} function $\varphi(s_0, t): \mathbb{R}^n \times \mathbb{R}_{\geq 0} \rightarrow \mathbb{R}^n$ is to describe the solution of the NNCS, which maps the initial state $s_0$ to the system state $\varphi(s_0, t)$ at time $t$ starting from $s_0$. We call a state $s'$ \textit{reachable} if there exist $s_0\in S$ and $t\in \mathbb{R}_{\geq 0}$ with $s'=\varphi(s_0,t)$. A reachable set $\mathcal{S}_r^T$ is a collection of all reachable states within a time range $T=\mathbb{R}_{\geq 0}$ given an initial space $\mathcal{S}_0 = \{s_0\}$, i.e., $\mathcal{S}_r^T = \{ \varphi(s_0, t), \,|\, s_0\in \mathcal{S} \wedge  t\in T \}$. Intuitively, once the reachable set $\mathcal{S}_r^T$ is non-overlapping with the unsafe sets $\mathcal{S}_u$, safety is guaranteed for such an NNCS throughout the time horizon $T$. 

\smallskip
\noindent
\textbf{POLAR-Express.} POLAR-Express~\cite{wang2023polar} is a reachability analysis tool for NNCS based on polynomial arithmetic, developed upon POLAR~\cite{huang2022polar}. It uses Bernstein polynomial interpolation to over-approximate the non-differentiable activation functions to enable layer-by-layer Taylor-Models (TMs) propagation for general feed-forward neural networks. The output over-approximation from the neural network is combined with Flow*~\cite{chen2015flowstar} for next-step reachable set computation. This process repeats with the previous reachable set result as the input set for the next step and thus rolls out the overall reachable set step by step within the entire time horizon. 
Moreover, to tighten the over-approximation, POLAR-Express stores the TM intervals symbolically with their linear transformation matrix and only evaluates the remainder interval at the end. This approach is called symbolic remainder, which reduces the accumulation of over-approximation error in TM by avoiding the wrapping effect in linear mappings. 

\subsection{Task Specification}

\begin{wrapfigure}{bR}{0.4\textwidth}
\vspace{-5em}
\begin{center}
\includegraphics[width=0.9\linewidth]{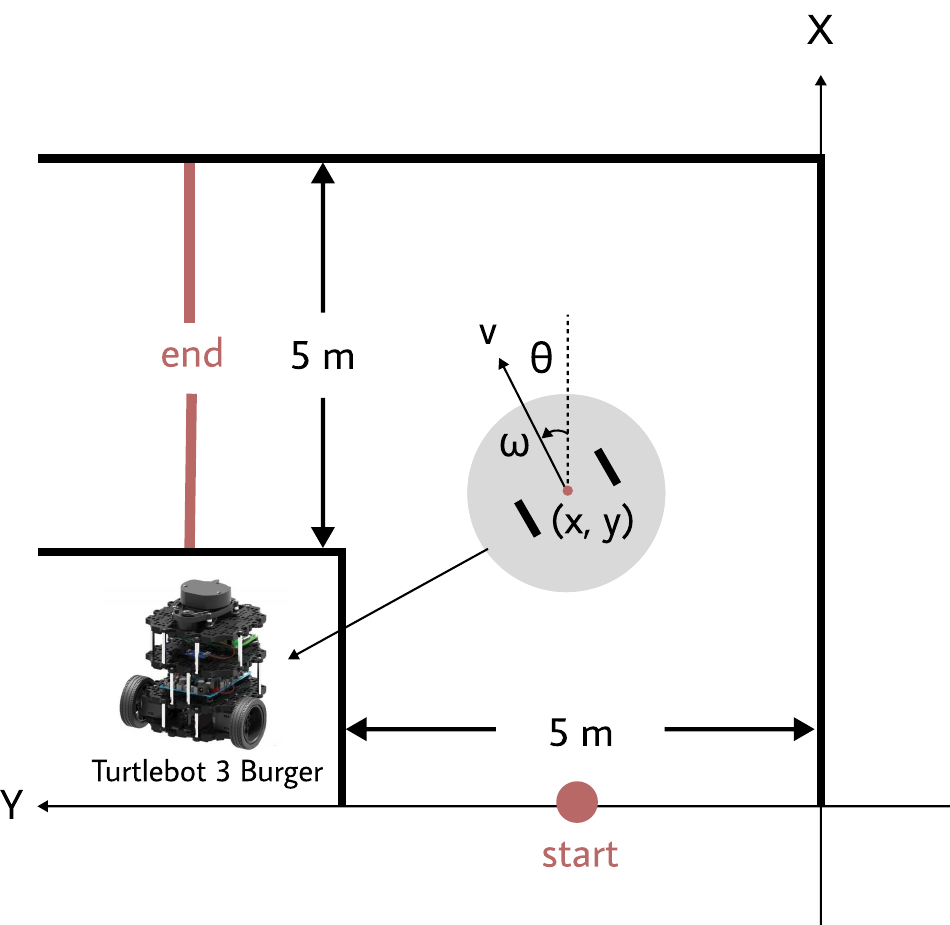}
\end{center}
\vspace{-2em}
\caption{Turtlebot Task Overview}
\vspace{-5em}
\label{fig:mapOverview}
\end{wrapfigure}

We control the Turtlebot 3 Burger (Details in Appendix~\ref{sec::turtlebot}) in the ROS2 Flatland simulation
to execute a left turn via an NN controller in a structured environment bounded by 5-meter walls (Fig.~\ref{fig:mapOverview}).
The Turtlebot is equipped with LiDAR sensing capabilities, enabling it to localize and detect obstacles within its surroundings. While the NN controller computes the desired speed and steering angle for the left turn, POLAR-Express runs in real-time to verify the controller's safety. To create dynamic and uncertain environments, we introduced random obstacles during navigation, which do not exist in the training phase of the NN controller. This scenario ensured that some of the NN control signals would be unsafe, thereby requiring POLAR-Express to capture unsafe maneuvers in real time and demanding a safe control adaptation strategy. 

\subsection{Runtime Verification (POLAR-Express) based Safe Control}

\begin{figure}[!htb]
    \centering
    \begin{subfigure}[b]{0.65\textwidth}
        \includegraphics[width=\textwidth]{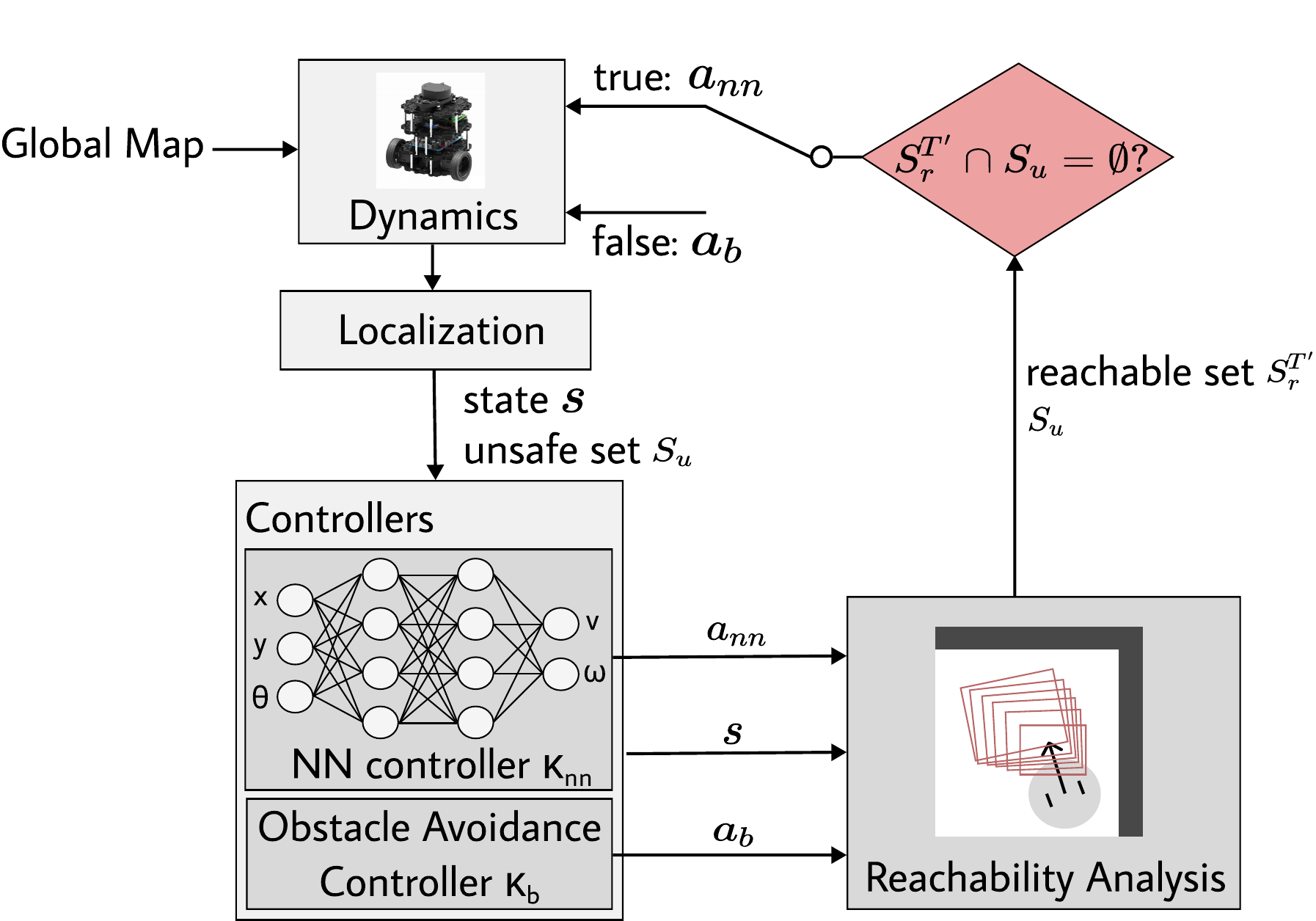}
        \caption{}
	\label{fig:controller_flowchart}
    \end{subfigure}
    \hfill
    \begin{subfigure}[b]{0.3\textwidth}
        \includegraphics[width=\textwidth]{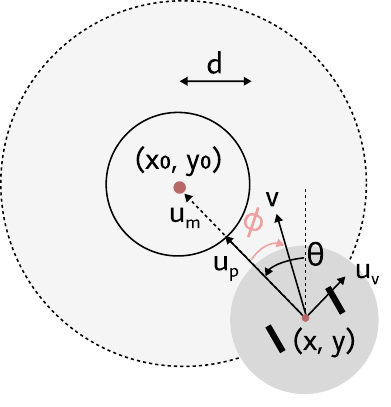}
        \caption{}
	\label{fig:geometricRelation}
    \end{subfigure}
    \caption{\textbf{a}. The case study framework. We use POLAR-Express to determine the switch between an NN controller and an obstacle avoidance controller. \textbf{b}. The obstacle avoidance controller design moves the Turtlebot counterclockwise while keeping a safe distance. 
    }
\end{figure}

Fig.~\ref{fig:controller_flowchart} outlines this case study's safe closed-loop control framework. We use POLAR-Express to compute reachable sets of NN controller $\kappa_{nn}$ for Turtlebot at runtime. In case of a potential collision, the Turtlebot is switched to a backup obstacle avoidance controller $\kappa_b$ for safety. We switch back to the NN controller if it is verified to be safe after the obstacle avoidance controller takes over. We introduce the details of each component in the following. 

\vspace{-1em}

\noindent
\textbf{Turtlebot Dynamics and localization.} We model the Turtlebot's dynamics as $\dot{x} = \cos(\theta)v$, $\dot{y} = \sin(\theta)v$, $\dot{\theta} = \omega$~\cite{Siwek2023}.
where $\theta$ is the orientation angle around x-axis and $(x, y)$ is the localized position, $s=[x, y, \theta]\in\mathcal{S}$. $a=[v,\omega]\in\mathcal{A}$ is the control input signal, representing linear velocity and angular velocity, generated by the controller.  Given a global map of the environment and the laser scan data from the LDS-01, the Turtlebot can localize its position by the Adaptive Monte-Carlo Localization (AMCL) approach implemented in the Nav2 package~\cite{automated_benchmark_nav2}.

\smallskip
\noindent
\textbf{NN Controller $\kappa_{nn}$.} We construct an NN controller $\kappa_{nn}: \mathbb{R}^3 \rightarrow \mathbb{R}^2$ with two hidden layers with a size of 64 neurons and ReLU activation functions. The controller takes the localized $(x, y, \theta)$ as input and outputs the linear velocity and angular velocity $(v, \omega)$ for the Turtlebot, i.e., $a_{nn}=(v, \omega)=\kappa_{nn}(x,y,\theta)$. To train the network, we collected 100 trajectories from expert demonstration data at 20 Hz in a simulation environment using the Nav2 goal package, moving the robot from a desired starting position to an end zone, and thus obtaining a dataset of $\{(x, y, \theta, v, \omega)\}$. We then train the NN controller via supervised learning to reduce an MSE loss as $\norm{\kappa_{nn}(x, y, \theta) - (v, \omega)}^2$. It is worth noting that there are no obstacles in the environment during offline training.

\label{sec:obs_avoid_controller}
\smallskip
\noindent
\textbf{Obstacle Avoidance Controller $\kappa_b$}. If the runtime verification result of $\kappa_{nn}$ is unsafe, we switch to the obstacle avoidance controller $\kappa_b$. 
Given the obstacle position, $\kappa_b$ move the Turtlebot around the obstacle counterclockwise while keeping a constant distance $d$ by adapting the algorithm in ~\cite{li2020switching}, as shown in Fig.~\ref{fig:geometricRelation}. 
Let $(x, y)$ be the robot's localized position and $(x_0, y_0)$ be the obstacle's center. The distance vector from the robot is $u_m = \begin{bmatrix} x_0 - x \\ y_0 - y \end{bmatrix}$. 
To maintain a safe distance $d$, we compute $u_{p} = u_m - \frac{u_m}{||u_m||} * d$, which points toward the obstacle if $\norm{u_m} \geq{d}$, and vice versa. 
To move in parallel with the obstacle, we rotate $u_p$ by 90 degrees:  where R=$\begin{bmatrix} 0, -1 \\ 1, 0 \end{bmatrix}$. Combining both components, the desired motion and angle for safe obstacle avoidance is $u = u_p + u_v$ and $\phi = \arctan(u_x, u_y)$, where $u_x$ and $u_y$ are the projections of $u$ onto $x$- and $y$-axes, respectively. Considering Turtlebot's physical limits (0.22m/s linear and 2.84 rad/s angular), we cap the desired steering velocity $v$ while setting it to $\norm{u}$. Similarly, we cap the desired steering angle $\omega$ while setting it to the difference between $\phi$ and current orientation $\theta$. The control input of $\kappa_b$ becomes $$a_b = (v, \omega) = (\min(0.22, ||u||), \min(2.84, \phi - \theta)) $$
It is important to note that $\kappa_b$ is a fallback mechanism to steer the robot to safety when $\kappa_{nn}$ is deemed unsafe by runtime verification. For the scope of this work, we assume that $\kappa_b$ is guaranteed to safely navigate the robot around the obstacle to a stable point where $\kappa_{nn}$ can resume control.

\smallskip
\noindent
\textbf{Obstacle Detection as Unsafe Regions $\mathcal{S}_u$.} We introduce obstacles of random size and location on the NNCS trajectory for online navigation. Note that the neural network does not have any knowledge of the random obstacle on the map. Rather, these obstacles can be detected and localized by the sensing ability of the Turtlebot at runtime. The location of these obstacles is treated as the unsafe region $\mathcal{S}_u$ for the safety verification of the neural network controlled Turtlebot by POLAR-Express. 


\smallskip
\noindent
\textbf{Controller Switching Logic.} As mentioned, Turtlebot can detect and localize the obstacles' locations as unsafe regions $\mathcal{S}_u$. At runtime, we use POLAR-Express to compute an over-approximation of reachable set for $\kappa_{nn}$ starting from the current state $s$ within a time horizon $T'$ as $\mathcal{S}_r^{T'}$. If $\mathcal{S}_u$ overlaps with $\mathcal{S}_r^{T'}$, this indicates a potential collision between Turtlebot under $\kappa_{nn}$ and the obstacle within time horizon $T'$, and therefore we switch to $\kappa_b$ producing $a_b$ for safety. While operating under $\kappa_b$, the robot continues to perform online reachability analysis for $\kappa_{nn}$. If $\mathcal{S}_u$ is no longer overlapping with $\mathcal{S}_r^{T'}$, i.e., $S_r^{T'} \cap S_u = \emptyset$, the robot switches back to $\kappa_{nn}$, as its control input is verified to be safe. This synergistic approach leverages $\kappa_{nn}$ for efficient task execution and relies on $\kappa_{b}$ and reachability analysis to guarantee safety in complex and cluttered environments; the switching logic can be carried entirely online.

\section{Experiments}
\begin{figure}[!htb]
    \centering
    \begin{subfigure}[b]{0.32\textwidth}
        \includegraphics[width=\textwidth]{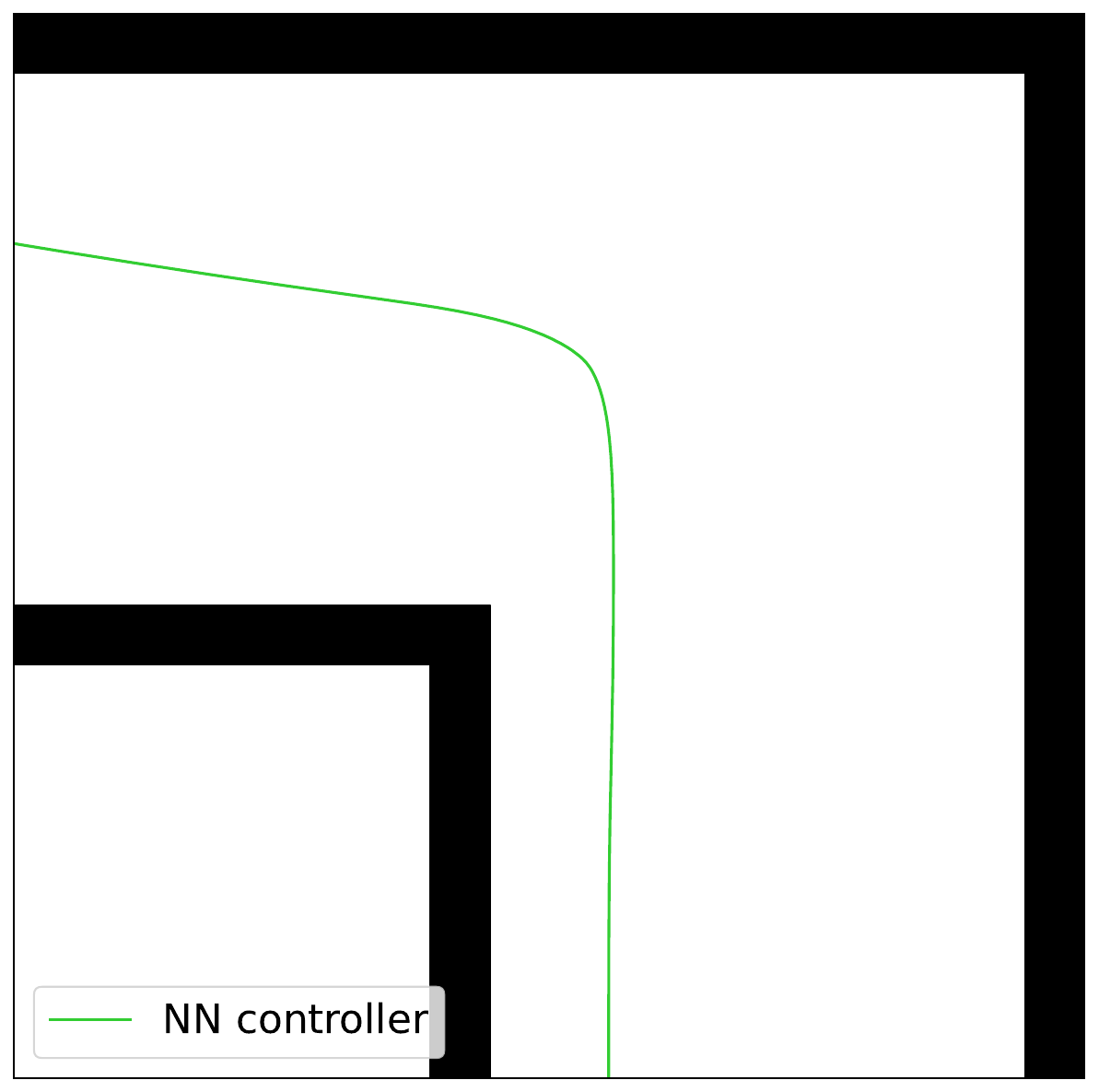}
        \caption{No Obstacle}
    \end{subfigure}
    \hfill
    \begin{subfigure}[b]{0.32\textwidth}
        \includegraphics[width=\textwidth]{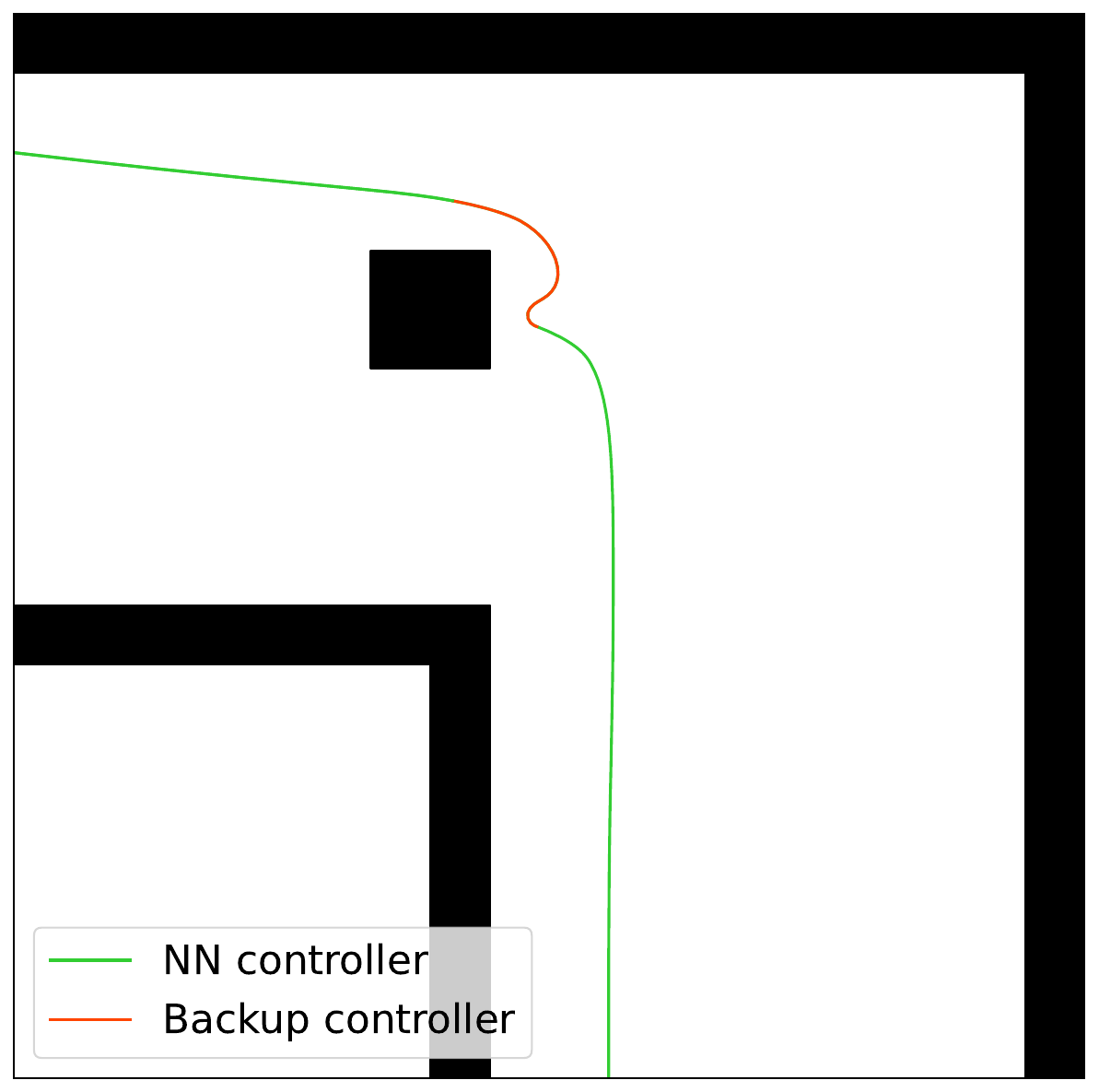}
        \caption{Single Obstacle}
    \end{subfigure}
    \hfill
    \begin{subfigure}[b]{0.32\textwidth}
        \includegraphics[width=\textwidth]{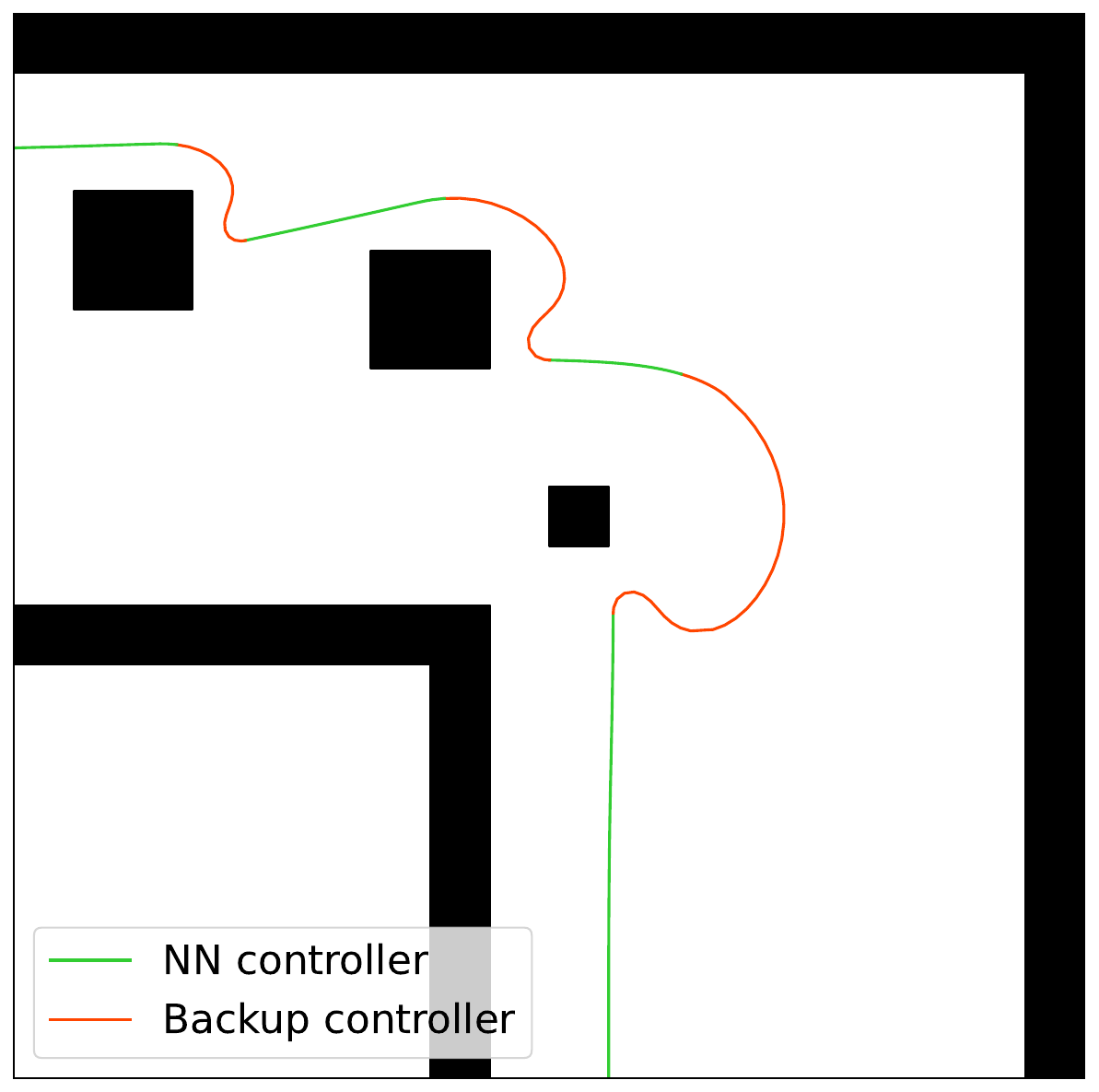}
        \caption{Multi-Obstacles}
    \end{subfigure}
    \caption{The navigation trajectory of Turtlebot with our runtime verification based control by $\kappa_{nn}$ (green) and $\kappa_b$ (red) for \textbf{a).} No obstacles, \textbf{b).} navigating around a single obstacle, and \textbf{c).} navigating through multiple obstacles. The connection points of green and red are the controller switching points.}
    \label{fig:overall_performance}
    
\end{figure}

The simulation was performed on a Dell XPS 15 with an i7 processor, performing reachability analysis every 0.2 seconds in the callback function of the ROS2 Flatland simulation. POLAR-Express can be customized by adjusting key hyper-parameters such as the degree of the Taylor Model (TM), the order of the Bernstein Polynomial approximation (BP), and the number of verification steps (please see~\cite{chen2013flow, wang2023polar} for more details of these hyper-parameters). By default, we assign the order of TM as 2 and the order of BP as 2 with 10 verification steps.
With the default parameters, our framework operates effectively in both single (Fig.~\ref{fig:overall_performance}b) and multiple obstacle avoidance scenarios (Fig.~\ref{fig:overall_performance}c). Our well-trained $\kappa_{nn}$ driving agent responds to obstacles detected with the reachable set computation by POLAR-Express timely and correctly activating the guarding condition, which then switches to the $\kappa_b$ controller, also shown in Fig.~\ref{fig:time_step_trajectories}. The $\kappa_{nn}$ resumes control once the agent steers around the obstacle and the reachable set no longer overlaps with unsafe areas (Fig.~\ref{fig:overall_performance}b, Fig.~\ref{fig:time_step_trajectories}). In the multiple-obstacles scenario, our runtime framework consistently manages several controller switches, ensuring safety throughout the operation (Fig.~\ref{fig:overall_performance}c). To comprehensively evaluate the case study, we then explore different parameter settings for POLAR-Express in different runtime scenarios, which may affect the tightness and computation efficiency of the reachable set.

\subsubsection{Verification Time Steps.} 
\begin{figure}[!htb]
    \centering
    \begin{subfigure}[b]{0.32\textwidth}
        \includegraphics[width=\textwidth]{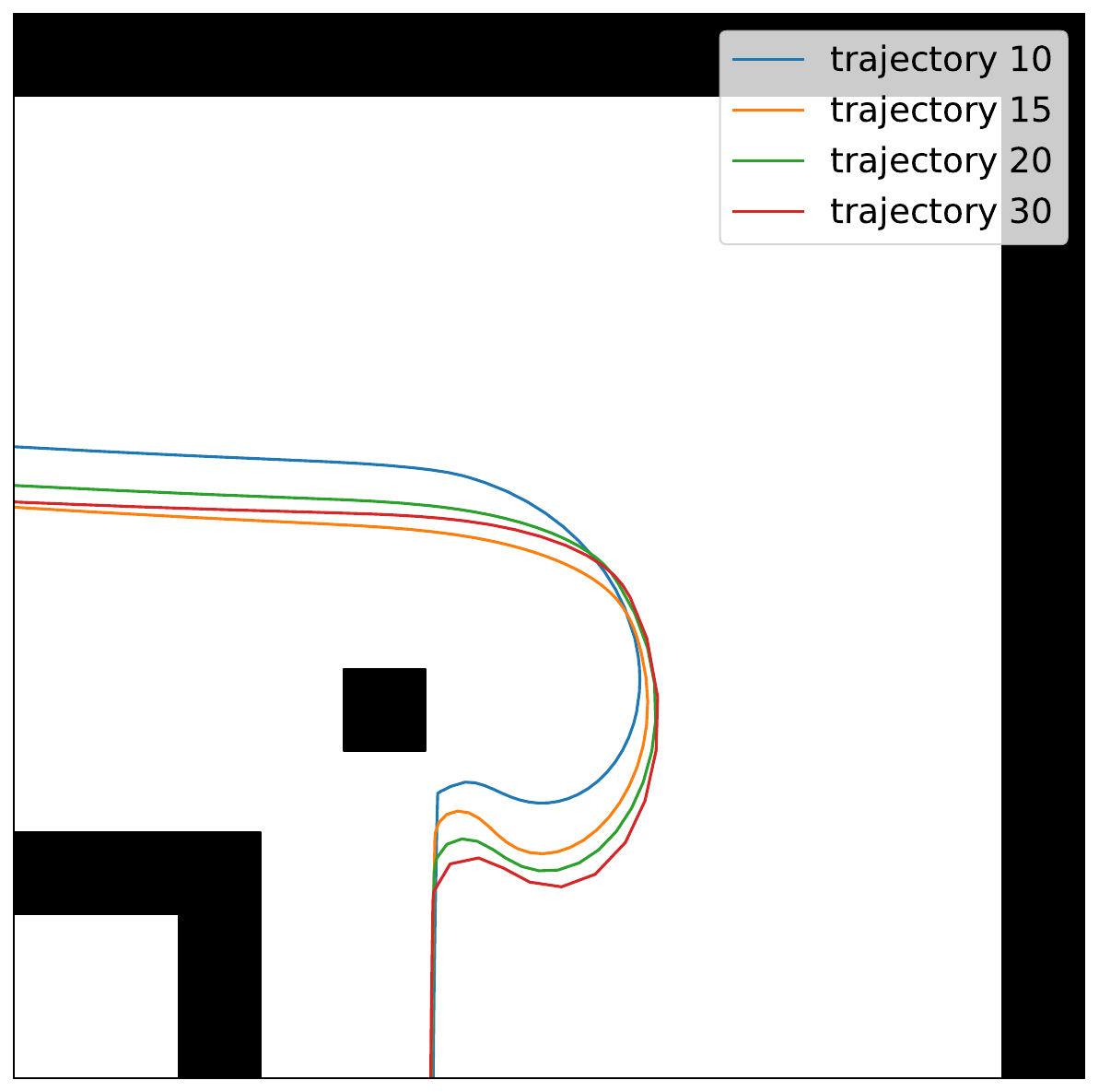}
        \caption{trajectories}
    \end{subfigure}
    \hfill
    \begin{subfigure}[b]{0.32\textwidth}
        \includegraphics[width=\textwidth]{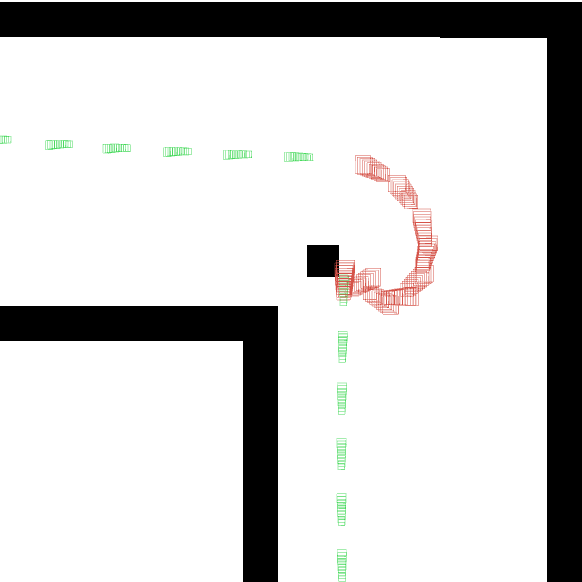}
        \caption{10 time steps}
    \end{subfigure}
    \begin{subfigure}[b]{0.32\textwidth}
        \includegraphics[width=\textwidth]{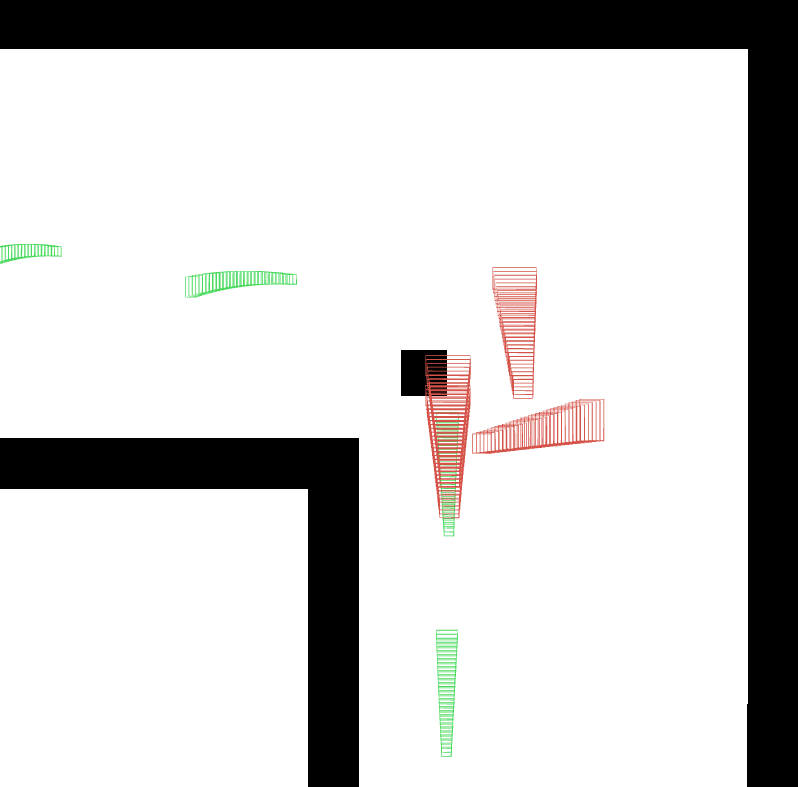}
        \caption{30 time steps}
    \end{subfigure}
    \caption{Trajectories and runtime reachable set visualization by POLAR-Express with varying verification time steps: green boxes and red boxes show runtime reachable sets using the NN and the obstacle avoidance controller, respectively.}
    \label{fig:time_step_trajectories}
\end{figure}

The verification time step determines the temporal horizon over which POLAR-Express computes the reachable set of the robot's future states. As observed in Fig.~\ref{fig:time_step_trajectories}a, longer verification time steps predict further and react to obstacles further ahead, while shorter steps react closer to obstacles. Although this predictive capability is desirable, increasing the verification time step introduces several drawbacks, as shown below.

Longer verification time steps increase the computational cost and may not satisfy the real-time verification requirement, as evidenced by the runtimes for different time steps in Table~\ref{table:table_runtime}. Fig.~\ref{fig:time_step_trajectories} demonstrates that we continue using $\kappa_{nn}$ if its runtime reachable set (green bounding boxes) does not overlap with the obstacle, and switch to $\kappa_b$ otherwise, indicated by the red runtime reachable set. The visualization shows that controllers with longer verification time steps, such as 30 (Fig. \ref{fig:time_step_trajectories}c), compute less frequently than those with shorter time steps, like 10 (Fig. \ref{fig:time_step_trajectories}b). Secondly, longer verification time steps can cause the controller to become more conservative and less task-critical, spending more time on obstacle avoidance and delaying task completion (Table~\ref{table:table_runtime}).
Lastly, longer verification steps may lead to an excessive accumulation of over-approximation error in the reachable set. This can result in an overly conservative evaluation of $\kappa_{nn}$'s safety, causing a premature switch to $\kappa_b$ and consequent performance degradation.
\begin{table}
    \centering
    \begin{tabular}{|c|c|c|c|c|c|}
      \hline
      Verification Steps & 10 & 15 & 20 & 30 \\ \hline
      Runtime (s) & 0.18 & 0.26 & 0.35 & 0.53 \\ 
      Task Total Time Usage(s) & 77.73 & 80.05 & 82.32 & 89.31 \\
      Obstacle Avoidance Controller Time Usage(s) & 20.97 & 22.37 & 25.19 & 28.99 \\
      Obstacle Avoidance Controller Utilization (\%) & 26.97 & 27.94 & 30.6 & 32.46 \\
      \hline
    \end{tabular}
    \caption{The verification time step vs. runtime (s)}
    \vspace{-2em}
    \label{table:table_runtime}
\end{table}

Overall, trajectory planning systems face a trade-off between computational complexity and safety considerations. Longer verification time steps ensure safer navigation by exploring more potential paths and identifying obstacles earlier, but this comes at the cost of increased computational time and data sparsity, potentially causing delayed verification decisions and reduced task criticality. Conversely, shorter verification time steps may be computationally more efficient but risk overlooking potential obstacles or failing to plan adequately. Striking the right balance between these factors is crucial for performance and safety.




\subsubsection{Timing of Runtime Verification with POLAR-Express.}

In this section, we evaluate the runtime performance of POLAR-Express by conducting experiments with different combinations of Taylor Model (TM) degrees and Bernstein Polynomial (BP) approximation orders, which determines the accuracy of NN approximation and dynamic systems propagation. Intuitively, higher order degrees of the polynomials within POLAR-Express provide more powerful and accurate approximations but come with more computation burden. 

\begin{wrapfigure}[15]{r}{0.5\textwidth}
    \begin{center}
    \vspace{-3em}
    \includegraphics[width=0.45\textwidth]{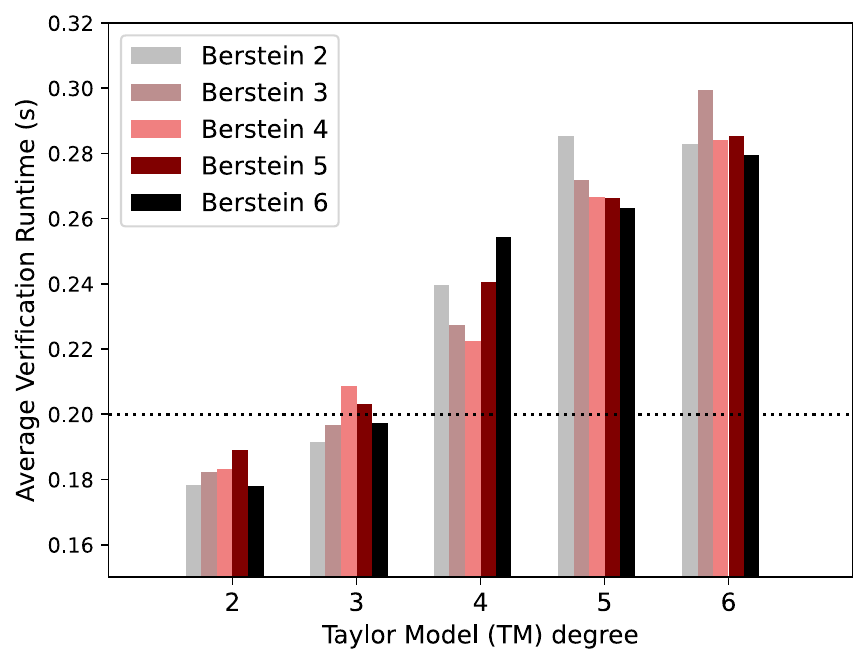}
    \end{center}
    \vspace{-1em}
    \caption{The 10-step verification runtime of POLAR-Express with different TM and BP orders.}
    \label{fig:degree_runtime}
\end{wrapfigure}

The bar graph in Fig. \ref{fig:degree_runtime} represents the runtime for various TM degrees and BP orders. Each runtime data is collected from the callback function and averaged for 10 trajectories. The POLAR-Express setup is fixed at 10 verification steps. In our evaluation, high TM degrees directly result in longer runtimes. We found that an increase in BP order does not drastically increase the time cost of the verification. Since the simulation is set to run the callback function every 0.2 seconds, only combinations with an average runtime of less than 0.2 seconds are considered valid for real-time performance. Based on this criterion, the valid combinations include TM degrees up to 3. 

 
\section{Conclusion and Future Work}

This paper presents a runtime verification case study where an autonomous Turtlebot, equipped with a neural network (NN) controller, navigates a structured environment using only LiDAR measurements and POLAR-Express for runtime reachability analysis. 
Our research can expand in several directions: 1) 
Adapting our framework to accommodate system uncertainties and stochastic policies is a potential area for further development. 2)  Incorporating scheduling techniques from the real-time systems for runtime verification could enhance system-level efficiency, where we can opportunistically call the verification engine only when it is necessary.
3) The switching logic of this case study is static and relatively simple, overlooking the fact that the obstacle avoidance controller could enter states that are not recoverable by the NN controllers. This could also be a future direction for improving this work.
\bibliographystyle{splncs04}
\bibliography{reference}

\newpage
\appendix
\section{Turtlebot Specification}
\label{sec::turtlebot}
To emulate real robot operations, we designed a robot testbed using the Flatland simulation environment. This setup replicates the dynamics of the Turtlebot 3 Burger, a differential wheeled robot equipped with two independently driven wheels and a LiDAR sensor~\cite{gross2020robot}. Its maximum translational and angular velocities are 0.22 m/s and 2.84 rad/s, respectively. It has a 360-degree Laser Distance Sensor (LDS-01) capable of scanning the environment at 300 rpm, with a distance range of 120 mm to 3600 mm and a sample rate of 1.8k Hz. Given the Turtlebot's LiDAR scanning distance range, we set up a simulation with 5-meter bounded walls (Fig.~\ref{fig:mapOverview}) to ensure the robot receives appropriate laser scan values for localization.

\end{document}